\documentclass{./styles/svproc}
\usepackage{graphicx,verbatim}
\usepackage{mathtools}  % amsmath with extensions
\usepackage{amssymb}  % (otherwise \mathbb does nothing)
\usepackage{caption}
\usepackage{subcaption}
\usepackage{booktabs}
\usepackage{multirow}
\usepackage{tikz}
\usetikzlibrary{calc}

\usepackage[acronym]{glossaries}
\newacronym{lvef}{EF}{Ejection Fraction}
\newacronym{es}{ES}{End-Systolic}
\newacronym{ed}{ED}{End-Diastolic}
\newacronym{a4c}{A4C}{Apical 4 Chamber}
\newacronym{a2c}{A2C}{Apical 2 Chamber}
\usepackage{url}

\begin{document}
\mainmatter              % start of a contribution
\title{InfoMotion: A Graph-Based Approach to Video Dataset Distillation for Echocardiography}
\titlerunning{InfoMotion}  % abbreviated title (for running head)
\author{Zhe Li \inst{1}\(^\ast\) \and Hadrien Reynaud\inst{2} \and
Alberto Gomez\inst{3} \and Bernhard Kainz\inst{1,2}}
\authorrunning{Zhe Li et al.} % abbreviated author list (for running head)
%\tocauthor{Ivar Ekeland, Roger Temam, Jeffrey Dean, David Grove}
%
\institute{Department AIBE, FAU Erlangen-Nürnberg, Erlangen, Germany, \\
\email{zhe.li@fau.de},\\
\and
Department of Computing, Imperial College London, London, UK,\\
\and 
Ultromics Ldt., Oxford, UK}

\maketitle              % typeset the title of the contribution

\begin{abstract}
Echocardiography plays a critical role in the diagnosis and monitoring of cardiovascular diseases as a non-invasive real-time assessment of cardiac structure and function. However, the growing scale of echocardiographic video data presents significant challenges in terms of storage, computation, and model training efficiency. Dataset distillation offers a promising solution by synthesizing a compact, informative subset of data that retains the key clinical features of the original dataset.
In this work, we propose a novel approach for distilling a compact synthetic echocardiographic video dataset. Our method leverages motion feature extraction to capture temporal dynamics, followed by class-wise graph construction and representative sample selection using the Infomap algorithm. This enables us to select a diverse and informative subset of synthetic videos that preserves the essential characteristics of the original dataset. We evaluate our approach on the EchoNet-Dynamic datasets and achieve a test accuracy of \(69.38\%\) using only \(25\) synthetic videos. These results demonstrate the effectiveness and scalability of our method for medical video dataset distillation.
\end{abstract}

\section{Introduction}

Ultrasound echocardiography is a widely used non-invasive imaging modality for real-time assessment of cardiac structure and function.
A key clinical metric derived from echocardiography is the Ejection Fraction (EF), which measures the proportion of blood pumped by the heart with each beat and serves as an important indicator of cardiac function and heart failure risk~\cite{huang2016accuracy}. EF is computed from the End-Diastolic Volume (EDV) and End-Systolic Volume (ESV) as $EF = \frac{EDV-ESV}{EDV} \times 100$.

Deep learning models have achieved strong performance in automated EF estimation, classification, and segmentation of echocardiographic data~\cite{he2023blinded,yang2023graphecho,gomez2025simplifying}, substantially improving cardiovascular diagnostics. However, progress is hindered by data scarcity and stringent restrictions on data sharing due to privacy protection, data ownership, licensing, and regulatory requirements. Generative Adversarial Networks (GANs)~\cite{tomar2021content} and diffusion models~\cite{nguyen2024training,reynaud2024echonet} have been explored to create synthetic datasets without protected health information. 
While synthetic data can help overcome privacy barriers, large-scale video generation also increases storage demands and training costs.
This motivates the need for dataset distillation to produce small, information-rich synthetic datasets that retain essential clinical features while reducing data volume and computational overhead.

Dataset distillation has been well studied in natural images via matching-based~\cite{zhao2020dataset,zhao2021dataset,zhao2023DM,cazenavette2022dataset} and generative-assisted~\cite{cazenavette2023generalizing,su2024d} methods. In medical imaging, applications remain limited to 2D modalities such as histopathology~\cite{Li_Image_MICCAI2024,cong2024dataset} and MRI~\cite{kanagavelu2024medsynth}. Video distillation is less explored due to the complexity of temporal dynamics. For natural videos, Wang et al.~\cite{wang2024dancing} proposed a two-stage framework combining image-based distillation with temporal modeling. 
However, dataset distillation for medical videos remains unexplored. To the best of our knowledge, this is the first work to develop a dataset distillation method for medical videos.
Ultrasound videos present unique challenges for distillation: the high structural and appearance similarity across frames and samples can limit diversity in the distilled set. Addressing these challenges requires methods that explicitly capture temporal motion patterns and select representative samples at the video level.

In this work, we propose a novel approach for medical video dataset distillation that extracts dynamic motion features to distinguish videos and selects representative samples from a large set of synthetic echocardiogram videos using the Infomap algorithm~\cite{blocker2023map}. The large synthetic set is generated by a pretrained latent video diffusion model~\cite{reynaud2024echonet}. Unlike previous work~\cite{Li_Image_MICCAI2024}, we extract motion features from each video to better capture dynamic content and address the challenge posed by similar anatomical structures. Each motion feature vector represents a video and serves as a node in a class-specific weighted graph, from which the Infomap algorithm identifies a small, diverse set of representative samples. We then train downstream models on this distilled synthetic set, achieving compelling classification results.

\noindent\textbf{Contributions: }
(1) We present the first dataset distillation method for medical videos, applied to ultrasound echocardiography. The method produces a compact synthetic dataset generated via latent diffusion models, initiating dataset distillation research in this domain.
(2) We introduce an efficient, class-wise video selection strategy based on motion features. By constructing a graph for each class and applying the Infomap algorithm~\cite{blocker2023map}, we identify representative samples while overcoming the high appearance similarity between echocardiogram videos.
(3) We validate our method on the EchoNet-Dynamic~\cite{ouyang2020video-based} and EchoNet-Synthetic~\cite{reynaud2024echonet} datasets, achieving a test accuracy of \(69.38\%\) using only \(25\) synthetic videos, demonstrating that models trained on a small distilled set can achieve high accuracy on real test data.
 
\noindent\textbf{Related works: } 
Dataset distillation has been extensively studied for natural images, including matching-based approaches~\cite{zhao2020dataset,zhao2021dataset,zhao2023DM,cazenavette2022dataset} and methods assisted by generative models~\cite{cazenavette2023generalizing,su2024d}. In the medical image domain, several works develop dataset distillation on histopathology~\cite{Li_Image_MICCAI2024,cong2024dataset} or MRI images~\cite{kanagavelu2024medsynth}. For natural videos, Wang et al.~\cite{wang2024dancing} proposed a two-stage framework that applies image-based matching methods before incorporating temporal dynamics.
Although distillation research for medical videos is lacking, echocardiogram video generation has progressed rapidly, from early GAN-based methods~\cite{reynaud2022d’artagnan} to diffusion models~\cite{reynaud2023feature-conditioned}, and most recently to latent video diffusion models~\cite{reynaud2023feature-conditioned,zhou2024heartbeat}, which offer high image fidelity with efficient training and inference.
Graph-based methods have also been applied in echocardiogram domain. Gao et al.~\cite{gao2021automated} added a graph regularizer for view classification, Mokhtari et al.\cite{mokhtari2022echognn} estimated EF from static frame-based graphs without modeling temporal relationships, and Yang et al.~\cite{yang2023graphecho} built per-frame graphs for chamber regions to reduce domain shift in video segmentation, recursively updating these graphs from the start to the end of each video to incorporate temporal information. 
In contrast, our method constructs class-level graphs, where each node represents a entire video.

%----------------------------------------------------------------------
%----------------------------------------------------------------------
\section{Method}
\begin{figure}[tb]
   \centering
      \includegraphics[width=0.9\linewidth]{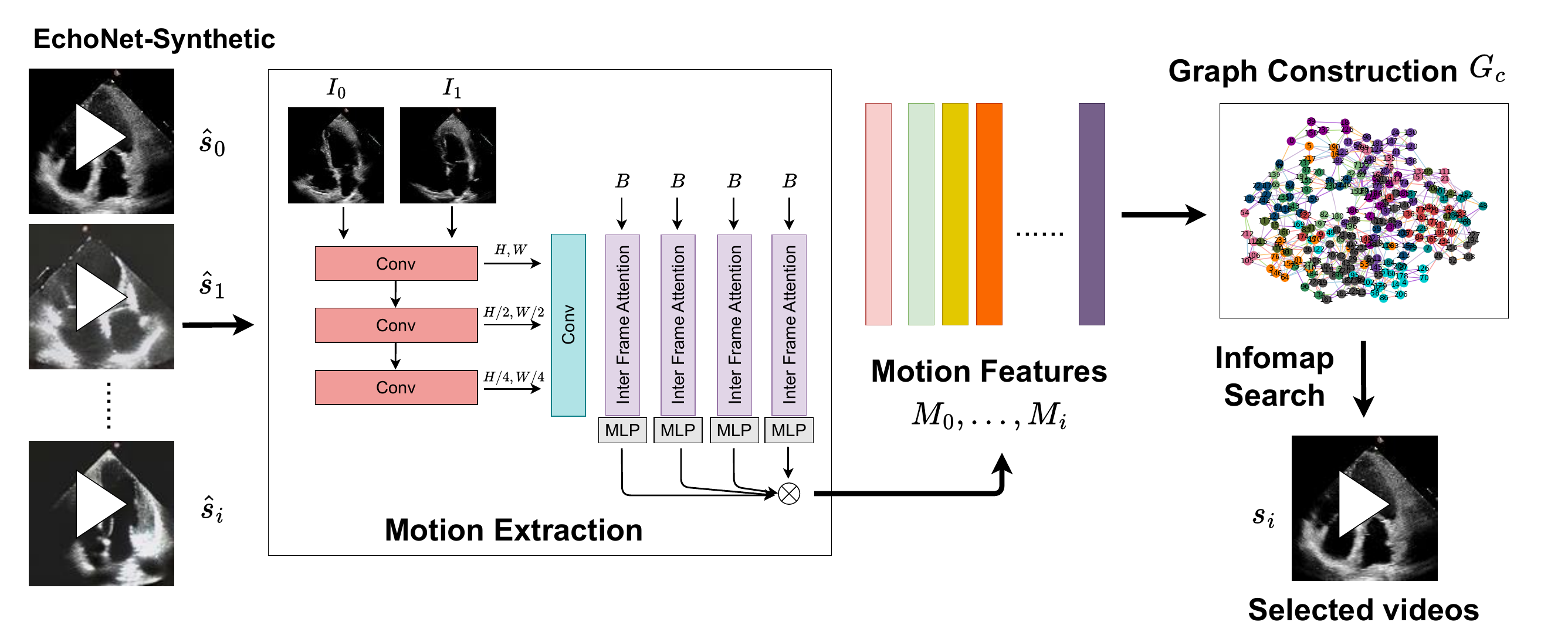}
   \caption{Overview of the proposed InfoMotion approach, which extracts motion features, constructs graphs, and applies a modified Infomap algorithm to identify representative synthetic videos.}
   \label{fig:overviewframework}
\vspace{-3mm}
\end{figure}

Fig.~\ref{fig:overviewframework} presents an overview of our proposed framework. We first extract motion-based features from echocardiogram videos and construct class-specific weighted graphs. The Infomap algorithm is then applied to identify a representative subset of synthetic videos. Our goal is to distill a small synthetic dataset that (1) captures the representative information of the real dataset and (2) achieves performance in downstream tasks comparable to training on the full real dataset.

\noindent\textbf{Problem definition: }\label{sec:problemdefinition}
Let the real ultrasound video dataset~\cite{ouyang2020video-based} be denoted as $\mathcal{T}=\{(v_i, EF_i)\}_{i=1}^{N}$, where $v_i\in \mathbb{R}^{T\times 3 \times H \times W}$ is a video with \(T\) frames, $N$ is total number of samples, and $EF_i$ is the corresponding \gls{lvef}. 
The objective is to synthesize a small set of videos $\mathcal{S}=~\{(s_i, y_i)\}_{i=1}^{N_s}$ with \(N_s = \text{VPC}\times C\), where \(C\) is the number of classes, $y_i \in \{0, ..., C-1\}$, and $N_s \ll N$.
Here, VPC (videos per class) controls the size of the distilled dataset; for example, VPC \(= 5\) indicates that only \(5\) synthetic videos are retained for each class.
We categorize the videos into $C=5$ classes based on their \gls{lvef} value following clinical definitions: Severe dysfunction (\(\text{EF}<30\%\)), Moderate dysfunction (\(30\% \leq \text{EF} < 39\%\)), Mild dysfunction (\(40\% \leq \text{EF} < 49\%\)), Normal (\(50\% \leq \text{EF} \leq 70\%\)), and Hyperdynamic (\(\text{EF}>70\%\)).

\noindent\textbf{Motion Extraction: }
In medical ultrasound videos, spatial structures are often highly similar across all frames and samples, making it difficult to categorize videos based solely on appearance features. In contrast, motion information in the temporal dimension plays a more critical role in distinguishing videos. Therefore, we adopt motion information between frames as the primary feature representation for ultrasound videos.
Many existing works employ 3D convolutional networks to jointly extract static appearance features and temporal dynamics. However, this implicit encoding of motion can limit the quality and discriminative power of the resulting video features. To address this, we employ the Inter-Frame Attention (IFA) approach~\cite{zhang2023extracting}, originally developed for video interpolation, which explicitly separates appearance and motion features. We train the IFA model on our ultrasound dataset to extract both appearance and motion features, but retain only the motion features to represent each video. This explicit motion representation more effectively captures temporal changes, as illustrated in the \textit{Motion Extraction} step of Fig.~\ref{fig:overviewframework}.

\begin{figure}[tb]
\begin{center}
\resizebox{0.8\columnwidth}{!}{%
     \begin{subfigure}[b]{0.48\linewidth}
         \parbox[][4.5cm][c]{\linewidth}{
         \includegraphics[width=\linewidth]{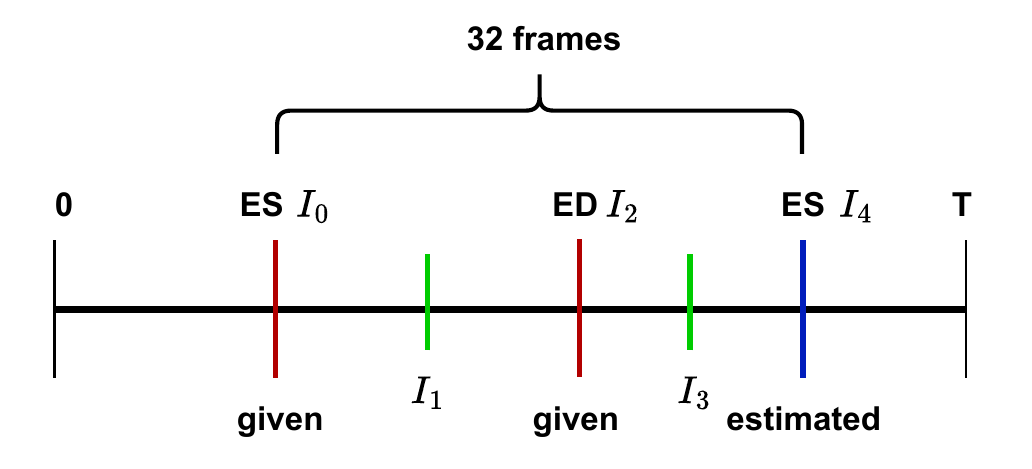} }
         \caption{Input frames}
     \end{subfigure}
     \begin{subfigure}[b]{0.48\linewidth}
         \includegraphics[width=\linewidth]{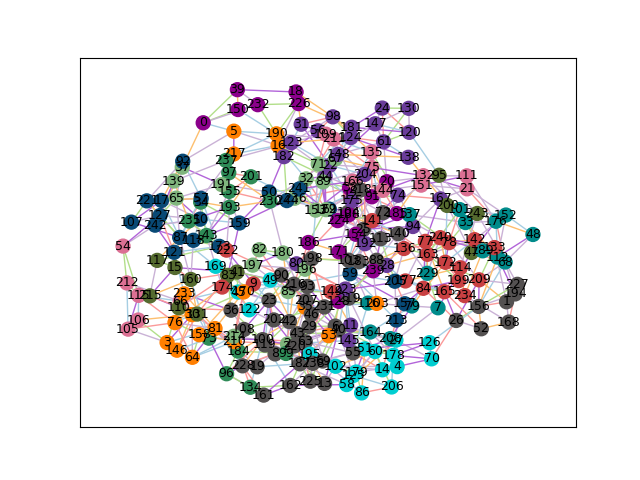}
         \caption{Graph example}     
     \end{subfigure}
}
\end{center}
\vspace{-3mm}
   \caption{(a) Estimation of the next ES frame index using the ground truth ES and ED frames. (b)Example graph for the \textit{Hyperdynamic} class generated by InfoMotion.}
\label{fig:motionextraction}
\vspace{-3mm}
\end{figure}

For a real video \( v_i \), we use a pair of consecutive End-Systolic (ES) and End-Diastolic (ED) frames as the input frames $I_0$ and $I_2$, as shown in Fig.~\ref{fig:motionextraction} (a). Our goal is to predict the middle frame $\hat{I}_1$, where the ground truth frame index is calculated by $\text{idx}(I_1) = \frac{\text{idx}(I_0)+\text{idx}(I_2)}{2}$. The predicted motion feature consists of two components: motion from $I_1$ to $I_0$ and motion from $I_1$ to $I_2$.
Since $I_0$ and $I_2$ cover approximately half a cardiac cycle, we also aim to capture motion over a full cycle. However, the subsequent ES frame \(I_4\) is not annotated. To estimate its index, we apply the clinically established assumption that, in a resting heart, diastole occupies approximately \(\frac{2}{3}\) of the cardiac cycle~\cite{hall2020guyton}, enabling us to infer the timing of the ED–ES transition relative to the annotated ED frame \(\text{idx}(I_4) = \text{idx}(I_2) + (\text{idx}(I_2) - \text{idx}(I_0)) * \frac{1}{2}\). 
Similarly, the middle frame between $I_2$ and $I_4$ is indexed as $\text{idx}(I_3) = \frac{\text{idx}(I_2)+\text{idx}(I_4)}{2}$.
We then input these frame pairs into the motion estimation module:

\begin{equation}\label{eq:estimatedmotion0}
    M_1 = Att(Conv(Concat(I_0,I_2))), \quad
    M_3 = Att(Conv(Concat(I_2,I_4))),
\end{equation}

where $M_1$ and $M_3$ are the extracted motion features, and $Conv$ denotes a convolutional layer, and $Att$ is the attention modules.
Using these motion features, we predict the middle frames \(\tilde{I}_1\) and \(\tilde{I}_3\), which are then refined via a lightweight refinement network $Rn$: $\hat{I}_1 = Rn(\tilde{I}_1)$ and $\hat{I}_3 = Rn(\tilde{I}_3)$. 
The objective loss combines a mean squared error (MSE) reconstruction loss between the refined frames and the ground truth, with a Laplacian loss~\cite{bojanowski2017optimizing} between the predicted frames and ground truth.
The Laplacian loss is defined as the L1 distance between the Laplacian pyramid representations of two images. The final loss is:

\begin{equation}\label{eq:loss}
    \mathcal{L} = \text{MSE}(\hat{I},I) + \lambda\mathcal{L}_{\text{Laplacian}}(\tilde{I}, I),
\end{equation}

where $\lambda$ is a hyperparameter to balance two losses.

\noindent\textbf{Graph construction: }
To identify representative videos that capture both essential information and the diversity of the dataset, we construct a weighted graph \( G_c =(M_c, E, w) \) for each class \( c \), where \( M_c=\{M_c^i\}_{i=1}^{N_c} \) is the set of nodes. Each node \(M_c^i\) corresponds to the motion feature vector of a video and $N_c$ denotes the total number of videos in class $c$. The edges \( E \) connect pairs of nodes, and the edge weights \( w \) are given by the Euclidean distance between their motion feature vectors.
We construct \(C\) separate graphs, one for each class, ensuring class-specific representation in the selection process.

\noindent\textbf{Infomap search: } 
Once the class-specific graphs are constructed, our objective is to identify representative nodes that capture the diversity and structure of each class. For this, we use the Infomap algorithm~\cite{blocker2023map}, which is well-suited for analyzing weighted graphs and uncovering meaningful community structures. Infomap applies the minimum description length (MDL) principle to model the movement of a random walker on the graph, partitioning it into communities that minimize the information required to describe this movement. This ability to detect informative, non-trivial community structures makes Infomap particularly effective for our motion-feature graphs.
%where similarities form dense, high-dimensional patterns.
After detecting communities, we uniformly select nodes with high modular centrality from each community. Modular centrality is a scalar metric that combines intra-community and inter-community influence scores~\cite{ghalmane2019centrality}, ensuring that selected nodes are both locally representative and globally influential. The resulting set of videos forms the distilled synthetic dataset used for downstream training.
Fig.~\ref{fig:motionextraction}(b) shows an example output of the Infomap search, where nodes of the same color belong to the same community.

%------------------------------------------------------------------

\begin{table}[tb]\setlength{\tabcolsep}{8pt}
\caption{Classification results of our approach for different VPC settings. 'w/o' indicates mapping EF values to classes without applying soft labels.}
\label{tab:infoclassification}
\begin{center}
\resizebox{0.9\columnwidth}{!}{
    \begin{tabular}{lclcc}
        \toprule
        Whole Real & All  & Dist. Real & VPC=5 & VPC=10 \\
        \midrule
        w/o Soft & 81.21 & Random & 61.47$_{\pm{8.0}}$ & 54.38$_{\pm{12.5}}$ \\
        & & Kmeans  & 62.44\(_{\pm{7.0}}\) & 62.51\(_{\pm{7.2}}\) \\
        & & InfoDist~\cite{Li_Image_MICCAI2024}  & 59.08$_{\pm{6.8}}$ & 61.27$_{\pm{5.7}}$ \\
        & & InfoMotion  & \textbf{65.15$_{\pm{3.3}}$} & \textbf{68.08$_{\pm{5.8}}$} \\
        \midrule
        Soft & 85.37 & InfoDist (soft)~\cite{Li_Image_MICCAI2024} & 68.94$_{\pm{11.9}}$ & 72.56$_{\pm{2.8}}$ \\
        && InfoMotion (soft) & \textbf{73.39$_{\pm{0.5}}$} & \textbf{73.86$_{\pm{1.7}}$} \\
        \midrule
        \midrule
        Whole Synth. & All & Dist. Synth. & VPC=5 & VPC=10 \\
        \midrule
        w/o Soft & 71.63 & Random & 62.73$_{\pm{11.2}}$ & 55.90$_{\pm{17.8}}$ \\
        && Kmeans & 68.19$_{\pm{3.4}}$ & 60.56$_{\pm{8.0}}$ \\
        && InfoDist~\cite{Li_Image_MICCAI2024} & 60.69$_{\pm{11.0}}$ & 65.67$_{\pm{5.7}}$ \\
        && InfoMotion & \textbf{69.38$_{\pm{4.0}}$} & \textbf{68.00$_{\pm{2.2}}$} \\
        \midrule
        Soft & 79.17 & InfoDist (soft)~\cite{Li_Image_MICCAI2024} & 67.27$_{\pm{10.8}}$ & 73.59$_{\pm{1.2}}$ \\
         && InfoMotion (soft) & \textbf{75.02$_{\pm{0.6}}$} &  \textbf{74.36$_{\pm{1.0}}$} \\
        \bottomrule
    \end{tabular}
}
\end{center}
\vspace{-5mm}
\end{table}

\begin{figure}[tb]
\begin{center}
\resizebox{0.8\columnwidth}{!}{%
     \begin{subfigure}[b]{0.48\linewidth}
        \parbox[][4cm][c]{\linewidth}{
        \includegraphics[width=\linewidth]{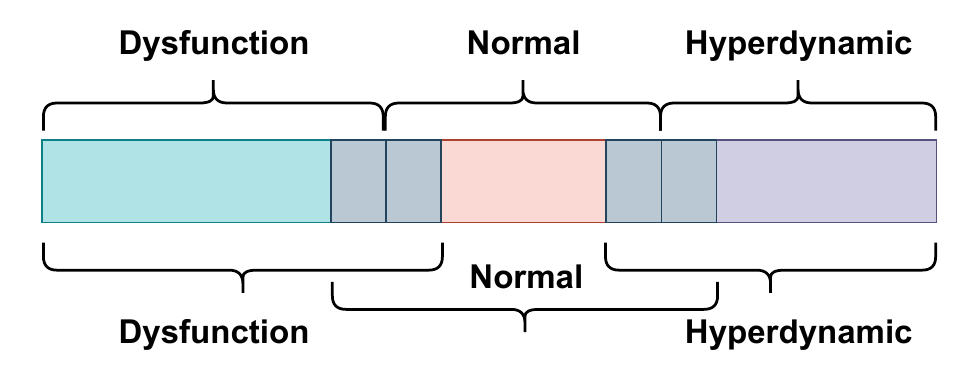}}
        \caption{Soft labels}
        \label{fig:infograph}
     \end{subfigure}    
     \begin{subfigure}{0.33\linewidth}
        \includegraphics[width=\linewidth]{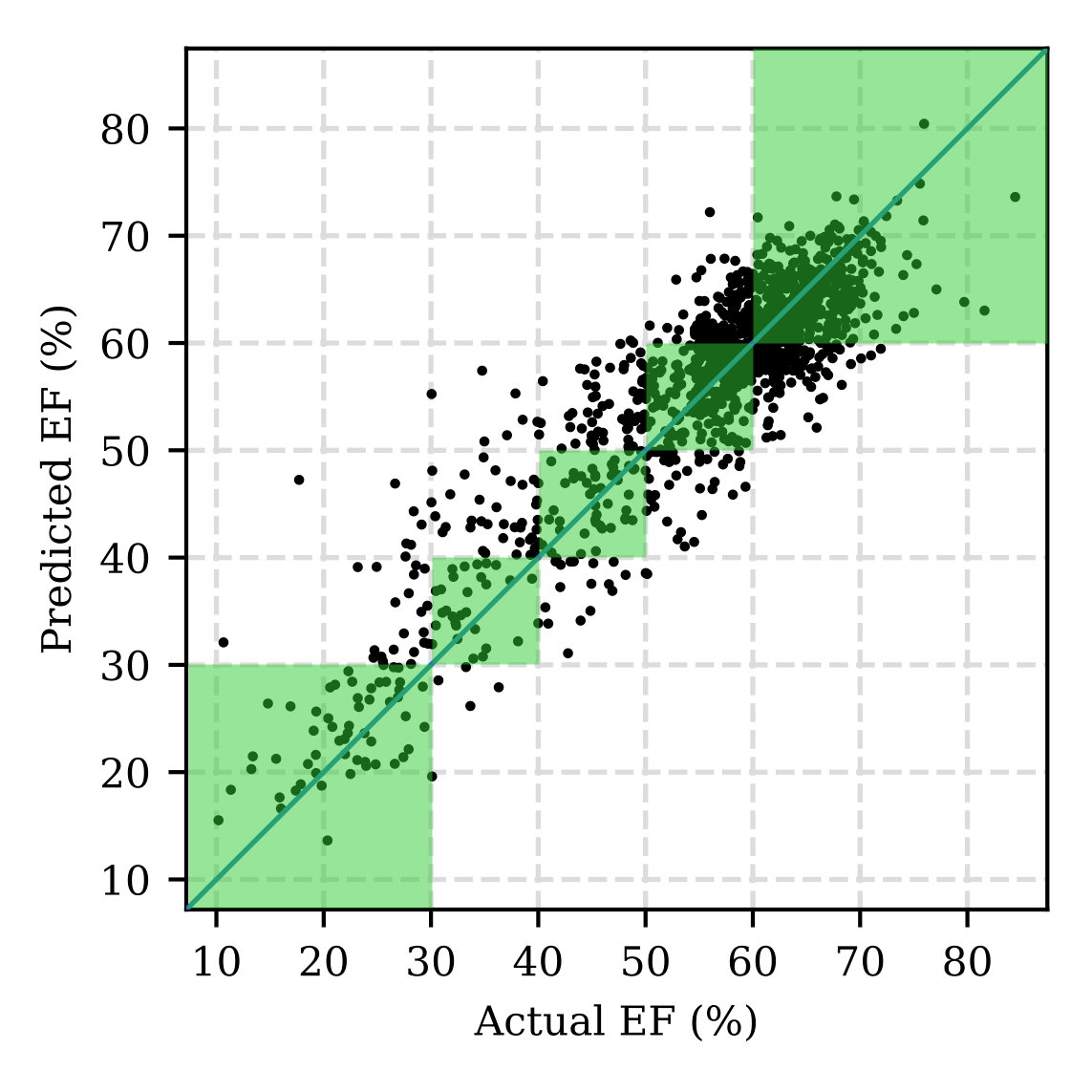}
        \caption{EF Predictions}
        \end{subfigure}
}
\end{center}
\vspace{-3mm}
   \caption{(a) Illustration of changes to class boundaries with soft labels. (b) Valid prediction regions in the soft setting, where green areas indicate predictions counted as correct in classification.}
\label{fig:motionsoftlabels}
\vspace{-3mm}
\end{figure}

\section{Experiments}

\noindent\textbf{Datasets: } 
We evaluate our approach on two cardiac ultrasound datasets: EchoNet-Dynamic~\cite{ouyang2020video-based} and EchoNet-Synthetic~\cite{reynaud2024echonet}.
EchoNet-Dynamic contains \(10,030\) \allowbreak echocardiogram videos with corresponding \gls{lvef} scores estimated by trained clinicians. The dataset is split into \(7465\) training, \(1288\) validation, and \(1277\) testing videos. Each video has a resolution of $112 \times 112$ pixels, a variable number of frames (\(28–1002\)). 
EchoNet-Synthetic consists of \(7500\) synthetic echocardiogram videos generated using a latent video diffusion model with \gls{lvef} scores sampled from a uniform distribution. The videos share the same $112 \times 112$ resolution as EchoNet-Dynamic, but have a fixed length of \(192\) frames, resulting in a total number of frames comparable to EchoNet-Dynamic. We re-label the synthetic videos using the regression model from~\cite{ouyang2020video-based} to obtain \gls{lvef} scores for training.

\noindent\textbf{Metrics: } 
We evaluate performance using test accuracy (ACC) and standard deviation. Each experiment setting is repeated \(5\) times, and the reported ACC is the average over these runs. Instead of training models directly for classification, we perform regression to predict the \gls{lvef} value and then map the predictions to class labels according to the predefined clinical \gls{lvef} ranges described in Section~\ref{sec:problemdefinition}. All metrics are computed on the real test set of EchoNet-Dynamic.

\noindent\textbf{Training: }
For motion extraction, we train the IFA model on the EchoNet-Dynamic dataset with a batch size of \(8\), a learning rate of \(0.0002\) using cosine scheduling and \(2000\) warm-up steps. The hyperparameter $\lambda$ in the loss function is set to \(0.5\).
During inference, we extract multi-scale motion features for each video with dimensions \([2, 512, 14, 14]\) and \([2, 1024, 7, 7]\). We average each feature map over the last two dimensions, flatten the remaining dimensions into 1D, and concatenate the resulting vectors to form a final motion feature vector of size \([1, 3072]\) for graph construction.
For classification, we train regression models on the selected small synthetic dataset using the Mean Squared Error (MSE) loss to predict \gls{lvef} values.

\noindent\textbf{Evaluation: } 
The classification results are presented in Table~\ref{tab:infoclassification}, obtained using the ResNet18 model. As this is the first study on dataset distillation for echocardiography videos, direct comparisons with existing methods are not available. Therefore, we construct our own baselines to enable fair comparisons across both selection strategies and feature extraction methods within a single results table. 
\texttt{Whole Real} denotes the baseline performance on the full EchoNet-Dynamic dataset, while \texttt{Dist.Real} refers to models trained on the distilled real dataset obtained by selecting videos from the real dataset. The same convention is applied to the EchoNet-Synthetic dataset, where \texttt{Whole Synth.} denotes the baseline performance on the full EchoNet-Synthetic dataset, and \texttt{Dist.Synth.} refers to we select videos from the full synthetic dataset and train models on the distilled synthetic dataset. 
In the \texttt{Random} setting, videos are selected randomly for each class, whereas \texttt{Kmeans} projects videos into a 1D embedding space by a pretrained ResNet18 model and selects them via K-means clustering. 
\texttt{InfoDist}~\cite{Li_Image_MICCAI2024}, originally proposed for histopathology images, is adapted and trained by us on the ultrasound dataset. The embeddings are extracted from a pretrained ResNet18 model, and video selection is performed using the Infomap algorithm. 
In contrast, our proposed \texttt{InfoMotion} method replaces appearance-based embeddings with motion features to construct the graph, enabling more effective selection of representative videos via the Infomap algorithm.

Given the high variance among clinicians in \gls{lvef} labeling near boundary, we introduce a soft evaluation metric (Fig.~\ref{fig:motionsoftlabels}) to reduce false positives and false negatives near class boundaries. This metric extends each class boundary by a small tolerance, allowing predictions within this margin to be counted as correct. This accounts for uncertainty the same way a clinical professional would.
In our experiments, we set the tolerance to \(2,\) \emph{e.g.}, the boundary between dysfunction and normal at \(50\) becomes \(0\sim 52\) for dysfunction and \(48\sim 72\) for normal.
With this soft threshold, \texttt{InfoMotion} achieves notable gains: in \texttt{Dist.Real} with VPC=\(10\), accuracy improves from \(68.08\) to \(73.86\); in \texttt{Dist.Synth.}, it reaches \(75.02\) with VPC=\(5\), using only \(25\) videos. These results approach the full real dataset baseline of \(81.21\). Moreover, \texttt{InfoMotion} shows greater stability, with lower standard deviation, \emph{e.g.} \(\pm{2.2}\) vs. \(\pm{17.8}\) for \texttt{Random} when VPC\(=10\).

\begin{figure}[tb]
    \centering
    \resizebox{0.6\columnwidth}{!}{%
        \begin{tabular}{ccc}
            \begin{subfigure}{0.32\linewidth}
                \centering
                \includegraphics[width=\linewidth]{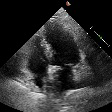}
                \caption{Real}
            \end{subfigure} &
            \begin{subfigure}{0.32\linewidth}
                \centering
                \includegraphics[width=\linewidth]{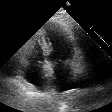}
                \caption{Echo-Synthetic}
            \end{subfigure} &
            \begin{subfigure}{0.32\linewidth}
                \centering
                \includegraphics[width=\linewidth]{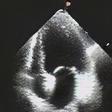}
                \caption{Latte}
            \end{subfigure}
        \end{tabular}%
    }
    \caption{Comparison of video quality between real and synthetic frames, shown for videos in class \(2\) label (\(40\%\leq\)EF \(\leq 49\%\)) from each dataset.}
    \label{fig:qualitativevideo}
\vspace{-3mm}
\end{figure}

\noindent\textbf{Ablation study: } 
To evaluate the robustness and generalizability of our InfoMotion approach, we compare its performance on synthetic datasets generated by different diffusion models. In addition to the EF-conditioned latent video diffusion model used for EchoNet-Synthetic, we test on videos generated by Latte~\cite{ma2024latte}, a class-conditioned latent diffusion model that has shown strong results on natural video generation. Using Latte, we synthesize \(3125\) videos, \(625\) per class across \(5\) classes. As shown in Table~\ref{comparisonlatte}, test accuracies are slightly lower but remain comparable to those on EchoNet-Synthetic. The gap is largely attributed to weaker class conditioning in Latte, which results in noticeably higher FID and FVD metrics and poorer qualitative outputs (Fig.~\ref{fig:qualitativevideo}). This comparison across diffusion models demonstrates that InfoMotion remains effective even under less favorable generative conditions.

\begin{table}[tb]\setlength{\tabcolsep}{6pt}
\caption{Comparison of results using different diffusion models.}
\label{comparisonlatte}
\begin{center}
\resizebox{0.95\columnwidth}{!}{%
\begin{tabular}{lccccc}
\toprule
Diffusion Models & FID$\downarrow$ & FVD$\downarrow$ & All Synth. & VPC=5 & VPC=10 \\
\midrule
EchoSyn~\cite{reynaud2024echonet} & 17.4 & 71.4 & 71.63\(\pm{1.18}\) & 69.38\(\pm{3.95}\) & 68.00\(\pm{2.17}\) \\
Latte~\cite{ma2024latte} & 60.7 & 473.1 & 70.73\(\pm{1.19}\) & 66.45\(\pm{3.82}\) & 61.10\(\pm{1.88}\) \\
\bottomrule
\end{tabular}
}
\end{center}
\vspace{-3mm}
\end{table}

\section{Conclusion}

We introduced a motion-based dataset distillation approach for echocardiography videos, leveraging motion features to better capture temporal dynamics, constructing class-wise graphs, and selecting representative samples via the Infomap algorithm. 
The pregenerated synthetic dataset EchoNet-Synthetic addresses both privacy concerns and computational costs while preserving essential clinical information. 
Experiments demonstrate that our approach achieves competitive accuracy using only a small number of synthetic videos, with lower variance than baseline methods. These results show that motion-guided selection can be an effective strategy for scalable and privacy-preserving medical video analysis.
While our work focuses on EF classification, the framework can be extended to segmentation or view classification, though additional challenges need to be addressed.

\paragraph{Acknowledgments:} HPC was provided by NHR@FAU - b143dc and b180dc. NHR is funded by federal and Bavarian authorities, with hardware partially funded by the DFG (440719683). H.R. was supported by Ultromics Ltd. and the UKRI CDT AI4Health (EP/S023283/1). We used Isambard-AI (AIRR)~\cite{mcintosh2024isambard}, operated by UoB and funded by UK DSIT via UKRI and STFC [ST/AIRR/I-A-I/1023]. Additional funding came from ERC MIA-NORMAL 101083647, DFG 513220538 and 512819079, and the state of Bavaria (HTA).

\bibliographystyle{./styles/bibtex/splncs03}
\bibliography{references}

@article{he2023blinded,
  title={Blinded, randomized trial of sonographer versus AI cardiac function assessment},
  author={He, Bryan and Kwan, Alan C and Cho, Jae Hyung and Yuan, Neal and Pollick, Charles and Shiota, Takahiro and Ebinger, Joseph and Bello, Natalie A and Wei, Janet and Josan, Kiranbir and others},
  journal={Nature},
  volume={616},
  number={7957},
  pages={520--524},
  year={2023},
  publisher={Nature Publishing Group UK London}
}

@book{gomez2025simplifying,
  title={Simplifying Medical Ultrasound: 5th International Workshop, ASMUS 2024, Held in Conjunction with MICCAI 2024, Marrakesh, Morocco, October 6, 2024, Proceedings},
  author={Gomez, Alberto and Khanal, Bishesh and King, Andrew and Namburete, Ana},
  volume={15186},
  year={2025},
  publisher={Springer Nature}
}

@incollection{mcintosh2024isambard,
  title={{Isambard-AI: a leadership-class supercomputer optimised specifically for Artificial Intelligence}},
  author={McIntosh-Smith, Simon and Alam, Sadaf and Woods, Christopher},
  booktitle={Proc. Cray User Group},
  pages={44--54},
  year={2024}
}

@inproceedings{cazenavette2022dataset,
	title        = {Dataset distillation by matching training trajectories},
	author       = {Cazenavette, George and Wang, Tongzhou and Torralba, Antonio and Efros, Alexei A and Zhu, Jun-Yan},
	year         = {2022},
	booktitle    = {CVPR'22},
	pages        = {4750--4759}
}

@inproceedings{cazenavette2023generalizing,
	title        = {Generalizing Dataset Distillation via Deep Generative Prior},
	author       = {Cazenavette, George and Wang, Tongzhou and Torralba, Antonio and Efros, Alexei A and Zhu, Jun-Yan},
	year         = {2023},
	booktitle    = {CVPR'23},
	pages        = {3739--3748}
}

@misc{ghalmane2019centrality,
	title        = {Centrality in modular networks. EPJ Data Sci 8 (1): 15},
	author       = {Ghalmane, Z and El Hassouni, M and Cherifi, C and Cherifi, H},
	year         = {2019}
}

@inproceedings{mokhtari2022echognn,
	title        = {{EchoGNN}: {Explainable} {Ejection} {Fraction} {Estimation} with {Graph} {Neural} {Networks}},
	author       = {Mokhtari, Masoud and Tsang, Teresa and Abolmaesumi, Purang and Liao, Renjie},
	year         = 2022,
	booktitle    = {Proc. MICCAI},
}

@article{ouyang2020video-based,
	title        = {Video-based {AI} for beat-to-beat assessment of cardiac function},
	author       = {Ouyang, David and He, Bryan and Ghorbani, Amirata and Yuan, Neal and Ebinger, Joseph and others},
	year         = 2020,
	month        = {April},
	journal      = {Nature},
	volume       = 580,
	pages        = {252--256}
}

@inproceedings{reynaud2022d’artagnan,
	title        = {D’{ARTAGNAN}: {Counterfactual} {Video} {Generation}},
	author       = {Reynaud, Hadrien and Vlontzos, Athanasios and Dombrowski, Mischa and Gilligan Lee, Ciarán and Beqiri, Arian and others},
	year         = 2022,
	booktitle    = {MICCAI},
}

@inproceedings{reynaud2023feature-conditioned,
	title        = {Feature-{Conditioned} {Cascaded} {Video} {Diffusion} {Models} for {Precise} {Echocardiogram} {Synthesis}},
	author       = {Reynaud, Hadrien and Qiao, Mengyun and Dombrowski, Mischa and Day, Thomas and Razavi, Reza and Gomez, Alberto and Leeson, Paul and Kainz, Bernhard},
	year         = 2023,
	booktitle    = {MICCAI},
	pages        = {142--152}
}

@string(CVPR= {IEEE Conf. Comput. Vis. Pattern Recog.})

@string(ICCV= {Int. Conf. Comput. Vis.})

@string(ICLR = {Int. Conf. Learn. Represent.})

@string(CVPR  = {CVPR})

@string(ICCV  = {ICCV})

@string(ICLR  = {ICLR})

@article{zhao2020dataset,
	title        = {Dataset condensation with gradient matching},
	author       = {Zhao, Bo and Mopuri, Konda Reddy and Bilen, Hakan},
	year         = {2020},
	journal      = {ICLR'20}
}

@inproceedings{zhao2021dataset,
	title        = {Dataset condensation with differentiable siamese augmentation},
	author       = {Zhao, Bo and Bilen, Hakan},
	year         = {2021},
	booktitle    = {ICML'21},
	pages        = {12674--12685},
	organization = {PMLR}
}

@inproceedings{reynaud2024echonet,
  title={Echonet-synthetic: Privacy-preserving video generation for safe medical data sharing},
  author={Reynaud, Hadrien and Meng, Qingjie and Dombrowski, Mischa and Ghosh, Arijit and Day, Thomas and Gomez, Alberto and Leeson, Paul and Kainz, Bernhard},
  booktitle={International Conference on Medical Image Computing and Computer-Assisted Intervention},
  pages={285--295},
  year={2024},
  organization={Springer}
}

@inproceedings{zhou2024heartbeat,
  title={Heartbeat: Towards controllable echocardiography video synthesis with multimodal conditions-guided diffusion models},
  author={Zhou, Xinrui and Huang, Yuhao and Xue, Wufeng and Dou, Haoran and Cheng, Jun and Zhou, Han and Ni, Dong},
  booktitle={International Conference on Medical Image Computing and Computer-Assisted Intervention},
  pages={361--371},
  year={2024},
  organization={Springer}
}

@inproceedings{tomar2021content,
  title={Content-preserving unpaired translation from simulated to realistic ultrasound images},
  author={Tomar, Devavrat and Zhang, Lin and Portenier, Tiziano and Goksel, Orcun},
  booktitle={Medical Image Computing and Computer Assisted Intervention--MICCAI 2021: 24th International Conference, Strasbourg, France, September 27--October 1, 2021, Proceedings, Part VIII 24},
  pages={659--669},
  year={2021},
  organization={Springer}
}

@inproceedings{nguyen2024training,
  title={Training-Free Condition Video Diffusion Models for Single Frame Spatial-Semantic Echocardiogram Synthesis},
  author={Nguyen, Van Phi and Luong Ha, Tri Nhan and Pham, Huy Hieu and Tran, Quoc Long},
  booktitle={International Conference on Medical Image Computing and Computer-Assisted Intervention},
  pages={670--680},
  year={2024},
  organization={Springer}
}

@InProceedings{Li_Image_MICCAI2024,
        author = { Li, Zhe and Kainz, Bernhard},
        title = { { Image Distillation for Safe Data Sharing in Histopathology } },
        booktitle = {proceedings of Medical Image Computing and Computer Assisted Intervention -- MICCAI 2024},
        year = {2024},
        publisher = {Springer Nature Switzerland},
        volume = {LNCS 15010},
        month = {October},
        page = {459 -- 469}
}

@article{cong2024dataset,
  title={Dataset distillation for histopathology image classification},
  author={Cong, Cong and Xuan, Shiyu and Liu, Sidong and Pagnucco, Maurice and Zhang, Shiliang and Song, Yang},
  journal={arXiv preprint arXiv:2408.09709},
  year={2024}
}

@article{blocker2023map,
  title={The Map Equation Goes Neural},
  author={Bl{\"o}cker, Christopher and Tan, Chester and Scholtes, Ingo},
  journal={preprint arXiv:2310.01144},
  year={2023}
}

@inproceedings{zhang2023extracting,
  title={Extracting motion and appearance via inter-frame attention for efficient video frame interpolation},
  author={Zhang, Guozhen and Zhu, Yuhan and Wang, Haonan and Chen, Youxin and Wu, Gangshan and Wang, Limin},
  booktitle={Proceedings of the IEEE/CVF Conference on Computer Vision and Pattern Recognition (CVPR)},
  pages={5682--5692},
  year={2023}
}

@article{ma2024latte,
  title={Latte: Latent diffusion transformer for video generation},
  author={Ma, Xin and Wang, Yaohui and Jia, Gengyun and Chen, Xinyuan and Liu, Ziwei and Li, Yuan-Fang and Chen, Cunjian and Qiao, Yu},
  journal={arXiv:2401.03048},
  year={2024}
}

@inproceedings{kanagavelu2024medsynth,
  title={MedSynth: Leveraging generative model for healthcare data sharing},
  author={Kanagavelu, Renuga and Walia, Madhav and Wang, Yuan and Fu, Huazhu and Wei, Qingsong and Liu, Yong and Goh, Rick Siow Mong},
  booktitle={MICCAI},
  pages={654--664},
  year={2024},
  organization={Springer}
}

@article{bojanowski2017optimizing,
  title={Optimizing the latent space of generative networks},
  author={Bojanowski, Piotr and Joulin, Armand and Lopez-Paz, David and Szlam, Arthur},
  journal={arXiv preprint arXiv:1707.05776},
  year={2017}
}

@inproceedings{wang2024dancing,
  title={Dancing with Still Images: Video Distillation via Static-Dynamic Disentanglement},
  author={Wang, Ziyu and Xu, Yue and Lu, Cewu and Li, Yong-Lu},
  booktitle={Proceedings of the IEEE/CVF Conference on Computer Vision and Pattern Recognition (CVPR)},
  pages={6296--6304},
  year={2024}
}

@article{zhao2023DM,
  title={Dataset Condensation with Distribution Matching},
  author={Zhao, Bo and Bilen, Hakan},
  journal={WACV'23},
  year={2023}
}

@inproceedings{su2024d,
  title={D\^{} 4: Dataset Distillation via Disentangled Diffusion Model},
  author={Su, Duo and Hou, Junjie and Gao, Weizhi and Tian, Yingjie and Tang, Bowen},
  booktitle={Proceedings of the IEEE/CVF Conference on Computer Vision and Pattern Recognition},
  pages={5809--5818},
  year={2024}
}

@article{gao2021automated,
  title={Automated recognition of ultrasound cardiac views based on deep learning with graph constraint},
  author={Gao, Yanhua and Zhu, Yuan and Liu, Bo and Hu, Yue and Yu, Gang and Guo, Youmin},
  journal={Diagnostics},
  volume={11},
  number={7},
  pages={1177},
  year={2021},
  publisher={MDPI}
}

@article{huang2016accuracy,
  title={Accuracy of left ventricular ejection fraction by contemporary multiple gated acquisition scanning in patients with cancer: comparison with cardiovascular magnetic resonance},
  author={Huang, Hans and Nijjar, Prabhjot S and Misialek, Jeffrey R and Blaes, Anne and Derrico, Nicholas P and Kazmirczak, Felipe and Klem, Igor and Farzaneh-Far, Afshin and Shenoy, Chetan},
  journal={Journal of Cardiovascular Magnetic Resonance},
  volume={19},
  number={1},
  pages={34},
  year={2016},
  publisher={Elsevier}
}

@inproceedings{yang2023graphecho,
  title={Graphecho: Graph-driven unsupervised domain adaptation for echocardiogram video segmentation},
  author={Yang, Jiewen and Ding, Xinpeng and Zheng, Ziyang and Xu, Xiaowei and Li, Xiaomeng},
  booktitle={ICCV},
  pages={11878--11887},
  year={2023}
}

@book{hall2020guyton,
  title={Guyton and Hall Textbook of Medical Physiology E-Book: Guyton and Hall Textbook of Medical Physiology E-Book},
  author={Hall, John E and Hall, Michael E},
  year={2020},
  publisher={Elsevier Health Sciences}
}

% \begin{thebibliography}{6}

% \end{thebibliography}
\end{document}